\documentclass{article}
\usepackage[final]{nips_2017}
\usepackage[utf8]{inputenc} % allow utf-8 input
\usepackage[T1]{fontenc}    % use 8-bit T1 fonts
\usepackage{hyperref}       % hyperlinks
\usepackage{url}            % simple URL typesetting
\usepackage{breakurl}
\usepackage{booktabs}       % professional-quality tables
\usepackage{amsfonts}       % blackboard math symbols
\usepackage{nicefrac}       % compact symbols for 1/2, etc.
\usepackage{microtype}      % microtypography
\usepackage{graphicx}
\usepackage{tcolorbox}
\usepackage{subfig}
\usepackage{wrapfig}
\usepackage{float}
\usepackage{xcolor} % to access the named colour LightGray
\definecolor{LightGray}{gray}{0.9}
\title{Optimizing the Songwriting Process: Genre-Based Lyric Generation Using Deep Learning Models}

\author{
  Tracy Cai \\
  Department of Computer Science\\
  Stanford University\\
  \texttt{cpcai@stanford.edu} \\
  \And
  Wilson Liang \\
  Department of Computer Science \\
  Stanford University\\
  \texttt{liangwil@stanford.edu} \\
  \And
  Donte Townes \\
  Department of Computer Science \\
  Stanford University\\
  \texttt{townes01@stanford.edu} \\
}

\begin{document}

% \begin{center}
% \includegraphics[width=3cm, height=0.7cm]{CS230}
% \end{center}

\maketitle

\begin{abstract}
The traditional songwriting process is rather complex and this is evident in the time it takes to produce lyrics that fit the genre and form comprehensive verses. Our project aims to simplify this process with deep learning techniques, thus optimizing the songwriting process and enabling an artist to hit their target audience by staying in genre. Using a dataset of 18,000 songs off Spotify, we developed a unique preprocessing format using tokens to parse lyrics into individual verses. These results were used to train a baseline pretrained seq2seq model, and a LSTM-based neural network models according to song genres. We found that generation yielded higher recall (ROUGE) in the baseline model, but similar precision (BLEU) for both models. Qualitatively, we found that many of the lyrical phrases generated by the original model were still comprehensible and discernible between which genres they fit into, despite not necessarily being the exact the same as the true lyrics. Overall, our results yielded that lyric generation can reasonably be sped up to produce genre-based lyrics and aid in hastening the songwriting process.
\end{abstract}

\section{Introduction}	
For many people in the music industry, crafting lyrics can be a difficult feat, especially when attempting to fit within the expected tone of a genre. When writing lyrics the inappropriate diction or lack of rhyming and flow can negatively impact the popularity of the music for the targeted audience. It can also be noted that during the songwriting process the lack of ideas when starting to write lyrics for a song can make the songwriting process longer and zap an artist of their creative flow. 

This project was deployed with the intention to aid in the crafting of song lyrics based upon an identified genre. The usage of our model would help to form ideas or lyrical motifs that tend to fit a genre in a quicker manner than traditional lyric generation, thus easing the songwriting process for many musical artists and further optimizing the songwriting process. Our interest was in seeing how relevant, recognizable, and lyrically sensible our model could be in producing 100-word lyrical measures from a dataset of Spotify songs with their lyrics. In this report, we shall review the multiple generative models we evaluated mainly being models framed after RNNs with LSTM layers, pre-trained gcp2 models, and other models found on GitHub. Knowing that we are using a generative model, we expect to have non-deterministic outputs of lyrics in each trial, but have tested output samples against metrics for text similarity for pairs of sequential phrases on 2 different metrics, ROUGE1-R and BLEU$^{[4]}$. 

Our algorithm simply takes in a short phrase or word and using our LSTM model trained on the the Spotify dataset, we output a 100 character phrase or set of phrases dependent on the model's interpretation. The output is currently set to a .csv file. 

\section{Related work}
\subsection{High-Level LSTM Research}
Regarding related projects, we have observed a similar project in CS230 by A. Apellanes and J. Wagner$^{[1]}$ who used transfer learning on an LSTM RNN. Some strengths in this project seems to be their use of transfer learning from a model pre-trained model of 200 Kanye West songs. The authors expanded the model to take 4000 songs while keeping the initial model weights. Uniquely, they made use of a mean squared loss function over Cross Entropy like most translation models seen. The most unique part of this project seemed to be the use of a rhyme index, where suffixes of generated words are used during post-processing to ensure their generated lines matched the rhyme scheme with their datatset. This seems to be a weakness when it comes to coherence of generated lines and flexibility. Other methods employed seemed to be the use of hyperparameter tuning with the panda process, restriction of syllables allowed per generated line, and usage of Rho to combat noisy gradient descent. This project is very similar to our own, albeit limited to generating lyrics to rap songs. Their model using LSTM seems to agree with our general model research like with the high-level T5 model from HuggingFace$^{[14]}$, however we deviate in the idea of using transfer learning as we wished to train our own weights and specialize our model.

\subsection{Pre-Processing Adaptations}
We saw that pre-processing of the training data seemed to be vital in how the text would be grouped and likely arranged by the model, thus projects like that proposed by M. Sidorov$^{[13]}$ had strengths in using a labeling system embedded into the training data. Other pre-processing techniques like the use of a syllabic indexing, encoding by word, and encoding by character yielded distinct results that affected coherency of generated sentences. The best approach seemed to be the use of tokens.
\subsection{Model Architectural Adaptations}
The overall research we did seemed to come to a consensus that lyric generation would have an architecture using LSTM cells, categorical cross-entropy loss, and Adam optimization. However, some modifications based on non-text LSTM generation projects like that of Z. Chen$^{[3]}$ et al. used RMS Prop as the optimizer as well as a Bidirectional LSTM layer at the model's start to expedite training loss which was successful for their  project. Other models like the non-text music generation model from K. Chou and R. Peng$^{[5]}$, use a standard 3-layer LSTM model but also during passing of the data to the fully-connected layers, a batch norm and 2 linear layers are added to produce a vector the name size as their input vector. They additionally use batch norm and ReLU activation between linear layers. Cross entropy loss is used, but so is a stochastic gradient descent optimizer$^{[12]}$. It is with this information that we wished to investigate the best optimizer to use as approaches to this commonly diverged.

\section{Dataset and Features}

The dataset we are using is a Kaggle dataset, “Audio Features and Lyrics of Spotify Songs”, featuring 18,000 songs from Spotify$^{[11]}$. Our dataset is 44MB and has features including song lyrics, song title, artist name, playlist genre, track popularity, the language of the lyrics, and other audio features of the tract like danceability, loudness, energy, etc. The dataset needed little cleaning besides filtering out only songs that had English lyrics. This left us with 15405 usable songs in our dataset. Upon further analysis, of the “playlist genre” of our song set, we saw that our dataset was rather evenly distributed by “playlist genre”, but some relative outliers were the genres of “pop”, which contains the highest number of songs at 3739 songs, and the genres with the least songs being “latin” and “edm” with 857 and 1758 songs, respectively. The average number of songs per genre is 2568. 
\begin{wrapfigure}[10]{r}{0.5\textwidth}
  \begin{center}
    \includegraphics[width = 5cm]{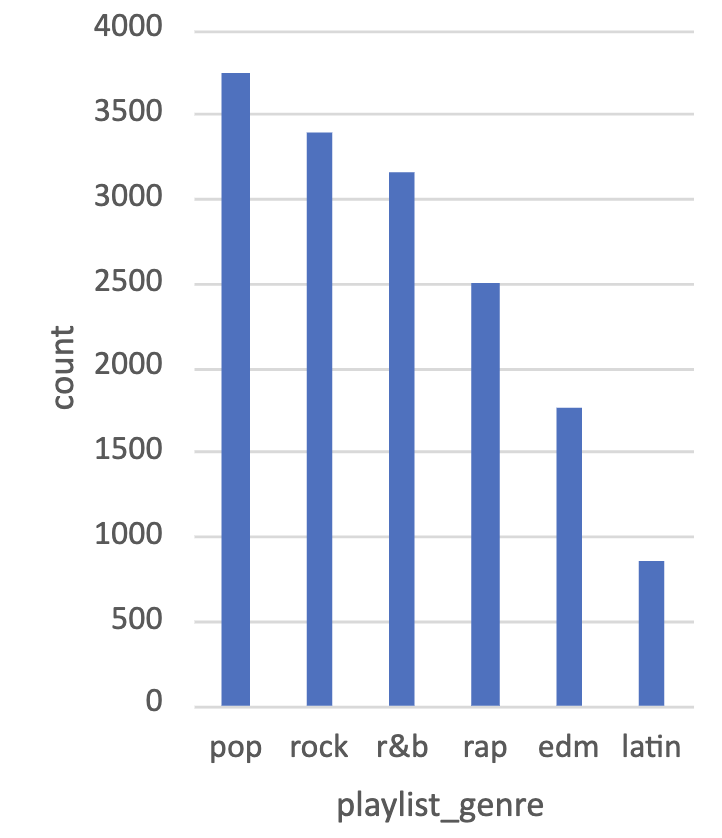}
  \end{center}
  \caption{Genre Breakdown of Dataset}
 \end{wrapfigure}
We preprocess our data by creating a corpus of all words found in the lyrics of our dataset and enumerating the unique words for the model to use. We did not explicitly divide our data into training, dev, and test sets because of the nature of a generative model. We split the lyrics of a song by verses (based on capitalization) and added a verse token '<V>' to indicate the delimiter. 

Lyrics Example Before Pre-processing:
\begin{tcolorbox}
When darkness falls, may it be That we should see the light When reaper calls, may it be That we walk straight and right When doubt returns, may it be That faith shall permeate our scars ...
\end{tcolorbox}
Post-Processing
\begin{tcolorbox}

['When darkness falls, may it be <V> That we should see the light <V> When reaper calls, may it be <V> That we walk straight and right <V> When doubt returns, may it be <V> That faith shall permeate our scars <V> ...
\end{tcolorbox}

Then we encode each word with a one-hot encoding according to a vocab dictionary created from the corpus$^{[6]}$. The encoded verses are the main features we will use to train our set to generate lyrics and the expected result would be the immediate lyrics that follow. 

\section{ Methods }
Our method was to develop 2 general models, an initial baseline model to get a generalization of what we may expect to output from a high-level, LSTM-based architecture and a custom model using PyTorch that would present a lower-level, LSTM architecture. 
\subsection{ Baseline Model}
The baseline model consists of a pre-trained, sequence-to-sequence model (seq2seq) from Hugging Face $^{[16]}$, specifically their AutoModelForSeq2SeqLM. This is a generic model class instantiated as a t5-small model with a sequence-to-sequence language modeling head. T5, specifically, is an encoder-decoder model which requires an input sequence and target sequence. It is trained using teacher forcing. Teacher forcing is a method for quickly and efficiently training recurrent neural network models that use the ground truth from a prior time step as input, thus previous data iterations use their subsequent counterpart to train on. We applied the T5 tokenizer (based on Google's unofficial SentencePiece tokenizer) to encode our data sets of verse sequences along with verse tokens. The baseline model also utilizes cross entropy loss in the model as this is standard for most language models. Its activation function is a softmax cross-entropy loss with masks.
We trained our baseline model with a training set of 681,442 english song entries and used a validation set with 170,360 entries. We then tested our model on 200 pairs of lyrical phrases. For our optimizer metric, we used an evaluation of the predicted and actual lyrics with Jaccard similarity.  
\[\frac{y_{prediction} \cap y_{actual}}{y_{prediction} \cup y_{actual}}\]

\subsection{ Final Model}
\begin{center}
\includegraphics[width=10cm]{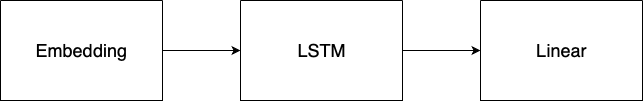}
\end{center}
For this model, we implemented a LSTM model using PyTorch which takes in prior sequential data to predict the next value. For each layer in the LSTM computes the following function on every element in the input sequence :
\begin{center}
\includegraphics[width=6cm]{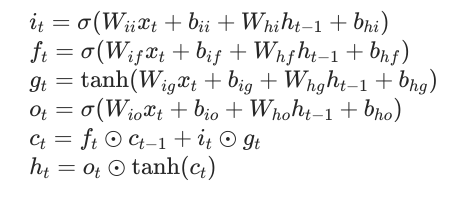}
\end{center}
We passed in our encoded lyrics using a sliding window approach with a sequence length of 4 (i.e given a 4-word sequence, the model predicts the next word). The sequence length of 4 was chosen to allow the model to learn from past words beyond the immediate previous word, but not too long as to impact training speed. The first layer of the model is the embedding layer, initialized according to our corpus vocab size. Next, we add three stacked LSTM layers, with input dimensions and hidden units set to 128 and a dropout rate of .2 for each layer. Finally we connect it to fully connected layer which outputs word vectors in dimensions according to the vocab size. We chose to stick with using Cross-Entropy Loss as our loss function as this was the standard consensus when investigating LSTM translation models. Cross-Entropy Loss in this case is used from pyTorch and specifically follows the formula without use of label smoothing, nor reduction: 
\begin{center}
\includegraphics[width=12cm]{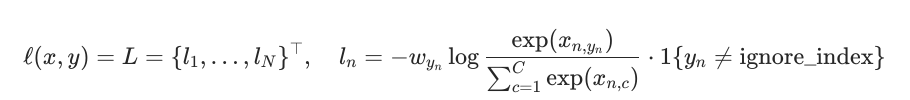}
\end{center}
Unlike our baseline model, however, we chose to use Adam$^{[9]}$ as a replacement optimizer for gradient descent as it tends to be efficient with large problems that utilize a lot of data$^{[17]}$.
The Adam optimizer does not require a large space, nor does it  need a lot of memory which helps us stay efficient in our gradient descent.
With our main model completed, we trained several new instances of this model using a subset of our data to produce three genre-based models: rock, pop, r\&b. Training each new model according to a specific genre would allow us to compare the effectiveness of specializing data compared to the larger model using all English songs from our dataset.
Formulas for ROUGE and BLEU metrics are seen in figure 3 (appendix).

\section{Experiments/Results/Discussion}
\subsection{Hyperparameter Tuning and Alterations}
After initial training of our baseline, we saw that we often had a very low accuracy, specifically, the average Jaccard similarity of the 200 prediction was 0.07743339232 with the best evaluation having a similarity of 0.8125. We ran 3 epochs on our baseline model, however and assumed that we would get better results with tweaks other model aspects, however, due to time-constraints, we could not investigate further hyperparameter tuning except to change the number of epochs to 2. This did appear to yield better results and less repetitions, nevertheless.
We focused most on modifying the tokenizer section of preprocessing as an independent variable. With different tokenizing schemes we deduced that we would get more comprehensible output by ensuring lines were properly split as well as punctuation and capitalization. This seemed to hold true as the form of our data returned clear, and notably similar to the form one would observe in lyrics.
\subsection{Quantitative Results}
We evaluated our models on two metrics, ROUGE1-R and BLEU.
We can see the results below:

\begin{center}  % Center the entire content
    \begin{minipage}{0.45\textwidth}  % Adjust the width of each minipage
        \centering  % Center the content within the minipage
        \begin{tabular}{|c@{\hskip 0.1cm}c@{\hskip 0.1cm}c@{\hskip 0.1cm}c|} 
        \hline 
        \multicolumn{4}{|c|}{Baseline Model Metrics} \\
        \hline
        Metric & Pop & R\&B & Rock \\
        \hline
        BLEU & 1.68e-227 & 1.22e-227 & 9.63e-228 \\ 
        \hline
        ROUGE & 2073.89 & 1235.07 & 1009.04 \\
        \hline
        \end{tabular}
    \end{minipage}
    \hspace{0.5cm}  % Space between the two tables
    \begin{minipage}{0.45\textwidth}
        \centering  % Center the content within the minipage
        \begin{tabular}{|c@{\hskip 0.1cm}c@{\hskip 0.1cm}c@{\hskip 0.1cm}c|} 
        \hline 
        \multicolumn{4}{|c|}{Original Model Metrics} \\
        \hline
        Metric & Pop & R\&B & Rock \\
        \hline
        BLEU & 2.17e-229 & 1.91e-229 & 2.21e-229 \\ 
        \hline
        ROUGE & 19.95 & 20.58 & 20.11 \\
        \hline
        \end{tabular}
    \end{minipage}
\end{center}
We can see that we had much higher ROUGE scores in our baseline, indicating a higher recall or how often training words appear in the model output, with the Pop model having the highest score of all. The original model has rather similar ROUGE scores leading us to interpret that we may have more originality in the original model than the baseline, but relatively even rates between genres.
For the BLEU metric, the baseline model seemed to score higher overall compared to our original model. This, however, leads us to believe that we have more precision in the baseline and there are more, but still few, words that appear in the model that appear in the training set. In turn, we had comparable BLEU scores over all genre models for the original model we made. This is an odd result, as with n grams trained with words, both scores should be high if comparing with the training set.
\subsection{Qualitative Results}
Below are some outputs from our original model that are noted to be highly comprehensible. We took note of punctuation being rather well-placed and even assessed by opinion the likeliness of each lyrical phrase to be used in a song of their respective genre. 

\begin{center}
    \textbf{Original model sample predictions:} \\    
\end{center}
\begin{center}  % Center the entire content
\begin{minipage}{0.45\textwidth}  % Create a minipage for the first table, adjust width
\centering  % Center the content inside the minipage
\begin{tabular}{|p{5cm}|}  % Fixed width for the column
 \hline
    \multicolumn{1}{|c|}{\textbf{Some R\&B Samples:}} \\    
 \hline
 Can we take it to the next level, baby, do you dare? \\ 
 \hline
Thinking of the fear I've had for so long \\
 \hline
Baby girl, we can do all the thangs you want to do \\
 \hline
I never knew there was a love like this before  \\
 \hline
I know you moved onto someone new \\
 [1ex] 
 \hline
\end{tabular}
\end{minipage}
\hspace{0.05\textwidth}  % Horizontal space between the two tables
\begin{minipage}{0.45\textwidth}  % Create a minipage for the second table
\centering  % Center the content inside the minipage
\begin{tabular}{|p{5cm}|}  % Fixed width for the column
 \hline
    \multicolumn{1}{|c|}{\textbf{Some Pop Samples:}} \\
 \hline
 You take me down, spin me around \\ 
 \hline
You promised me you'd be around \\
 \hline
I walk a little faster in the school hallway \\
 \hline
What else can we do when we're feelin' low?  \\
 \hline
Lights fill the streets, spreading so much cheer \\
 [1ex] 
 \hline
\end{tabular}
\end{minipage}
\end{center}

\begin{center}
\begin{tabular}{|c|} 
 \hline
    \multicolumn{1}{|c|}{\textbf{Some Rock Samples:}} \\
 \hline
 Love, like a road that never ends \\ 
 \hline
How've you been, have you changed your style? \\
 \hline
Trying to forget but i won't let go \\
 \hline
I've been waiting for you \\
 \hline
The radio station plays his latest song \\
 [1ex] 
 \hline
\end{tabular}
\end{center}

\section{Conclusion/Future Work }
Overall, we were successful in creating a rather simplistic LSTM model to generate song lyrics using a 3 layer LSTM model with Cross Entropy Loss, Adam optimization, and only 3 epochs of training. A large amount of our project focused on the best preprocessing technique to achieve a comprehensive output. However further implementations have been proposed to optimize the model and training. Some things that remain to be tested are hyperparameter tuning methods like creating multiple models where we change the batch size, number of LSTMs in a layer, number of layers, and perhaps type of layers. Another adaptation would be during evaluation. By feeding our model's output into a song genre classification model, we could assess what outputs are more fitting to their genre over others and tweak our model's weights according to what produces lyrics with the best mean fit. Finally, we could also tweak our preprocessing method by dividing our corpus by syllables or phonetics to achieve rhyme and rhythmic outputs.
\section{Appendix}
\subsection{Additional Figures}
\begin{figure}[H]%
    \centering
    \includegraphics[width=10cm]{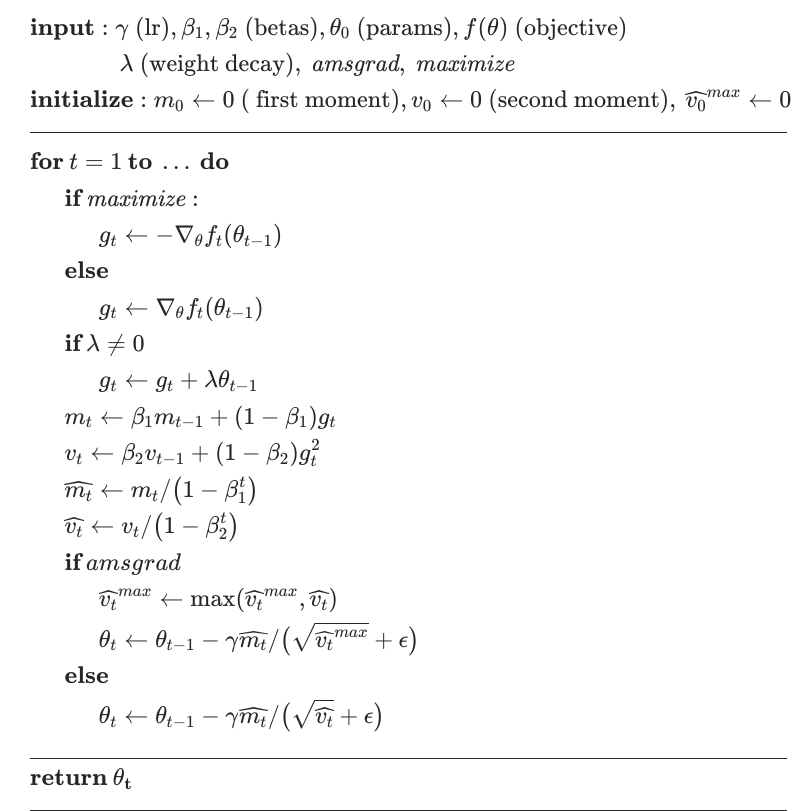}
    \caption{Adam Optimizer Formula}%
    \label{fig:example}%
    \centering
    \subfloat[\centering label 1]{{\includegraphics[width=3cm]{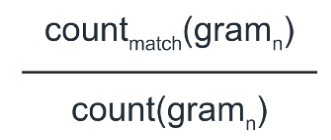}}}%
    \qquad
    \subfloat[\centering label 2]{{\includegraphics[width=5cm]{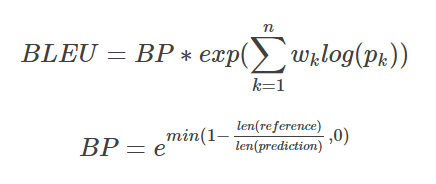}}}%
    \caption{ROUGE Formula (left), BLEU formula (right)}%
    \label{fig:example}%
\end{figure}

\end{document}